# Unsupervised particle sorting for high-resolution single-particle cryo-EM

Ye Zhou[1], Amit Moscovich[2], Tamir Bendory[3] and Alberto Bartesaghi[1,4,5]

[1] Department of Computer Science, Duke University, Durham, USA
[2] Program in Applied and Computational Mathematics, Princeton University, USA
[3] School of Electrical Engineering, Tel Aviv University, Tel Aviv, Israel
[4] Department of Biochemistry, Duke University School of Medicine, Durham, USA
[5] Department of Electrical and Computer Engineering, Duke University, Durham, USA

**Abstract**

Single-particle cryo-Electron Microscopy (EM) has become a popular technique for determining the structure of challenging biomolecules that are inaccessible to other technologies. Recent advances in automation, both in data collection and data processing, have significantly lowered the barrier for non-expert users to successfully execute the structure determination workflow. Many critical data processing steps, however, still require expert user intervention in order to converge to the correct high-resolution structure. In particular, strategies to identify homogeneous populations of particles rely heavily on subjective criteria that are not always consistent or reproducible among different users. Here, we explore the use of unsupervised strategies for particle sorting that are compatible with the autonomous operation of the image processing pipeline. More specifically, we show that particles can be successfully sorted based on a simple statistical model for the distribution of scores assigned during refinement. This represents an important step towards the development of automated workflows for protein structure determination using single-particle cryo-EM.

## 1. Introduction

Recent technological advances in cryo-electron microscopy (EM) have sparked a revolution in structural biology by enabling the study of challenging protein complexes that were inaccessible using X-ray crystallography or nuclear magnetic resonance (NMR) techniques [1], [2]. The introduction of direct electron detectors together with the development of effective computational image analysis tools for single-particle analysis (SPA) have led to a dramatic increase in the number of high-resolution structures of biologically important and diverse targets deposited in the Electron Microscopy Data Bank (EMDB) [3]. Some examples include large dynamic assemblies [4], [5], membrane transport proteins [6]–[9], G-protein coupled receptors (GPCRs) [10]–[13], and macromolecular complexes bound to small molecules [14]–[16]. The resolutions achieved in some cases rival those obtained by X-ray crystallography and NMR and are high enough to reveal features that are key for drug design, converting cryo-EM into a powerful method for protein structure determination.

At the many recently established cryo-EM facilities worldwide, raw data is collected in a fully automated manner taking advantage of unsupervised protocols for data acquisition. The increased access to cryo-EM infrastructure combined with the availability of better tools for data analysis, has significantly facilitated the traditionally tedious and time-consuming process of determining three-dimensional structures from cryo-EM samples. The computational structure determination process, however, remains an inexact art that requires significant levels of expert-user input in order to convert the large volume of noisy image data into high-resolution structures. The amount of user involvement varies depending on the level of difficulty of each sample, with lower molecular weight and highly dynamic or conformationally heterogeneous targets requiring the highest level of supervision. These technical challenges naturally pose barriers to the rapid advancement of the field because they slow down the overall structure determination process and can potentially introduce bias and subjectivity. Moreover, the imminent availability of larger and faster direct electron detectors will increase the present rates of data production even further, exacerbating the need to develop effective computational strategies that can handle the sheer amount of data produced at modern cryo-EM facilities.

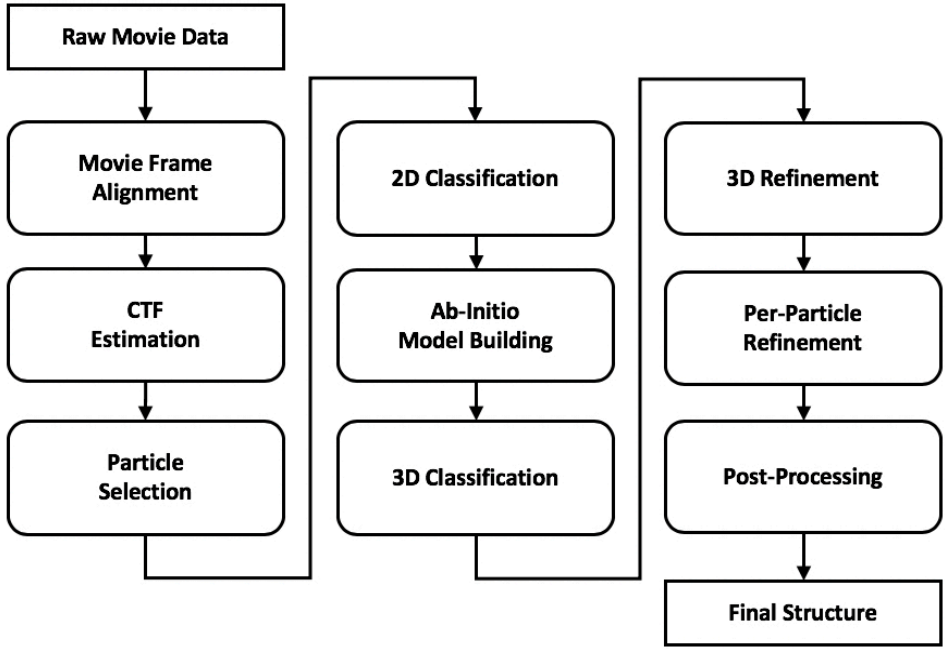

Figure 1. **Single particle cryo-EM image processing pipeline**. Flowchart diagram showing steps in image analysis required to convert raw movie data into high-resolution structures by single-particle cryo-EM. Raw movie data is first aligned to correct for beam induced motion and subjected to estimation of the Contrast Transfer Function (CTF). Individual molecular images are detected using particle selection, extracted and subjected to 2D heterogeneity analysis by image classification. Particles contributing to homogeneous subsets are used for ab-initio model building followed by further image classification in 3D. Homogeneous classes are then subjected to further 3D and per-particle refinement, to improve the resolution, and a final post-processing step is applied to obtain the final structure.

*1.1 Data processing pipeline for single-particle cryo-EM*

Converting single-particle cryo-EM image data into high-resolution structures requires solving an inverse problem where a large set of noisy two-dimensional projections is converted into a three-dimensional cryo-EM density map. In practice, this problem is solved using effective experimental strategies that follow a canonical sequence of data analysis steps, **Figure 1**. In the era of direct electron detectors, the raw data consists of sequences of dose-fractionated frames acquired on different areas of a cryo-EM grid. This movie data is first aligned in order to compensate for stage and beam-induced movements that occur during exposure of the sample to the electron beam [17]. Different computational strategies can be used to estimate these movements, including approaches for global [18]–[21], semi-local or patch-based [22], [23], and local or per-particle drift estimation [24]–[26]. After frame alignment, drift-corrected frame averages are produced and used for the processing downstream. The contrast transfer function (CTF) of the microscope for each movie is then estimated [27], [28], followed by identification of individual molecular images in each micrograph using *particle picking* strategies [29]–[33]. Boxes at the selected particle positions are then extracted and subjected to 2D image classification in order to eliminate false positives and identify homogeneous subsets of particles that can be used for *ab-initio* model determination [32], [34]–[39]. The resolution of this initial model is improved during the 3D refinement step [32], [34], [37], [40], followed by per-particle drift and CTF refinement to further improve map resolution [24], [26]. Finally, a post-processing or sharpening step is applied to the resulting map to compensate for the decay of amplitudes in frequency space, facilitating visual interpretation and fitting of atomic models into cryo-EM density maps [41], [42]. In the case of heterogeneous samples where a discrete number of conformations are present in the data, one or more rounds of 3D classification may be required in order to separate particle images into structurally homogeneous populations. These particle subsets can then be refined independently in order to produce high-resolution reconstructions for each conformation. More recently, several approaches have been proposed to tackle the important problem of continuous heterogeneity in cryo-EM samples, see [43] for a review. For further details on the computational pipeline for single-particle cryo-EM we refer the reader to [44] for a recent survey.

*1.2 Advances and challenges in automation*

Modern cryo-EM facilities routinely use automatic strategies for data collection as implemented in several proprietary and open-source data-acquisition suites [45], [46]. These tools have significantly increased the throughput of data collection by taking advantage of hardware improvements in modern electron microscopes and by automating tasks that typically rely on user input for their operation. Similarly, data processing packages for SPA have greatly facilitated the execution of many steps in the data analysis pipeline, including the implementation of robust strategies for movie-frame alignment, CTF estimation, particle selection, *ab-initio* model building, 2D/3D classification, 3D refinement, per-particle refinement, and post-processing [32], [34]–[36], [47]. Nevertheless, the overall computational structure determination process still requires significant levels of user involvement in order to successfully convert the noisy image data into high-resolution structures. This dependency on subjective user input not only slowdowns the structure determination process but can also compromise its consistency and reproducibility. One of the major barriers to automation in single-particle cryo-EM has been the need to manually identify homogeneous populations of particles that can yield high-quality structures. The most commonly used strategy to sort particle images consists of partitioning the set of molecular images into homogeneous subsets using consecutive rounds of 2D and 3D image classification, **Figure 2**. The success of this procedure, however, relies on a number of arbitrary decisions made by the user, including: a) selecting the appropriate number of clusters or classes in which to partition the data, b) choosing the correct subset of classes to keep or discard after each clustering step, and c) selecting the correct number and combination of 2D/3D classification rounds in order to successfully sort the particle images. In practice, many of these decisions are made according to heuristic principles or anecdotal evidence learned from extensive trial-and-error experimentation, and as such are difficult to reproduce, time consuming and more importantly, inconsistent with operation of the data processing pipeline in high-throughput mode.

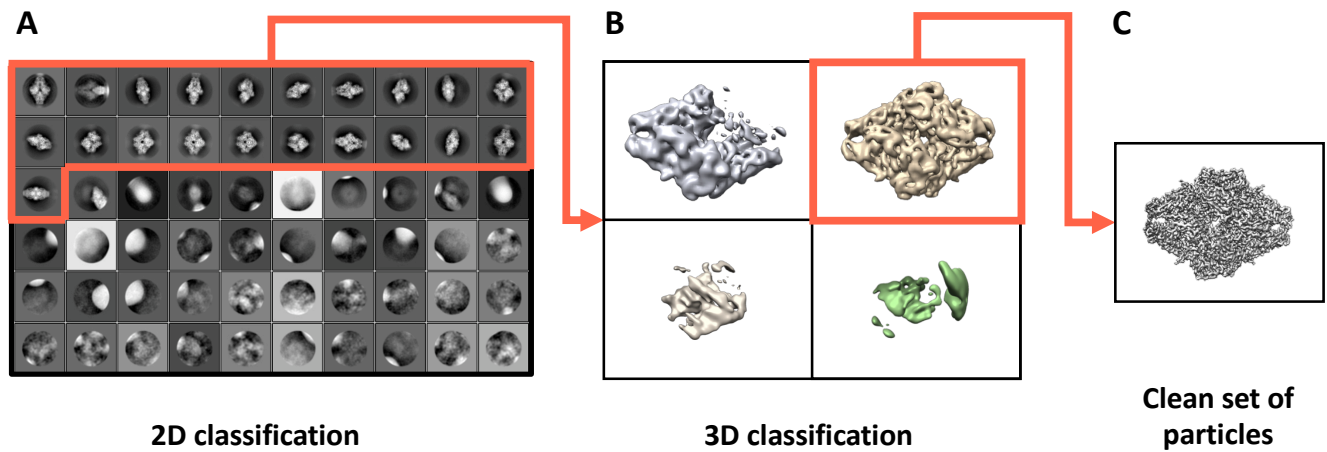

*Figure 2. **Supervised particle sorting based on 2D/3D image classification**. **A**) 2D classes obtained from the beta-galactosidase dataset (EMPIAR-10061) showing examples of homogeneous 2D classes separated by the image classification procedure and manually selected by a user (highlighted in orange). **B**) The subset of particles contributing to the classes selected in A was further classified into four classes using 3D classification. The most homogeneous class (highlighted in orange) was selected for further processing. **C**) High-resolution reconstruction obtained from particles contributing to the 3D class selected in B. The total number of classes used during 2D/3D classification and the selection of homogeneous classes in each round, both involve subjective user decisions that determine the success of the sorting procedure.*

Within this context, we seek to find ways to reduce the amount of user interaction needed during the single-particle cryo-EM structure determination process, by exploring the use of unsupervised strategies for image sorting. Availability of automated workflows for data processing will not only speed-up structure determination, but also improve its reproducibility and consistency. Moreover, even in challenging cases where automatic processing is unsuccessful, unsupervised image analysis can be combined with interactive processing in order to increase efficiency by allowing expert users to focus their efforts on the more challenging steps that are more relevant to study the biology. Advances in automation in data analysis will eventually allow the cryo-EM pipeline to operate in high-throughput mode, putting the technology on a path to establishing itself as an effective technique for high-resolution structure determination.

## 2. Methods

The fundamental premise of the standard tools for SPA is that experimental particle images correspond to projection views of the same underlying 3D object. The ability of these methods to identify homogeneous subsets of images which can be combined in 3D to produce a high-resolution reconstruction is therefore critical to their success. In practice, this is typically achieved through a two-stage procedure where 1) the location of putative particles is first identified within each micrograph (particle picking), followed by 2) application of sorting algorithms to identify

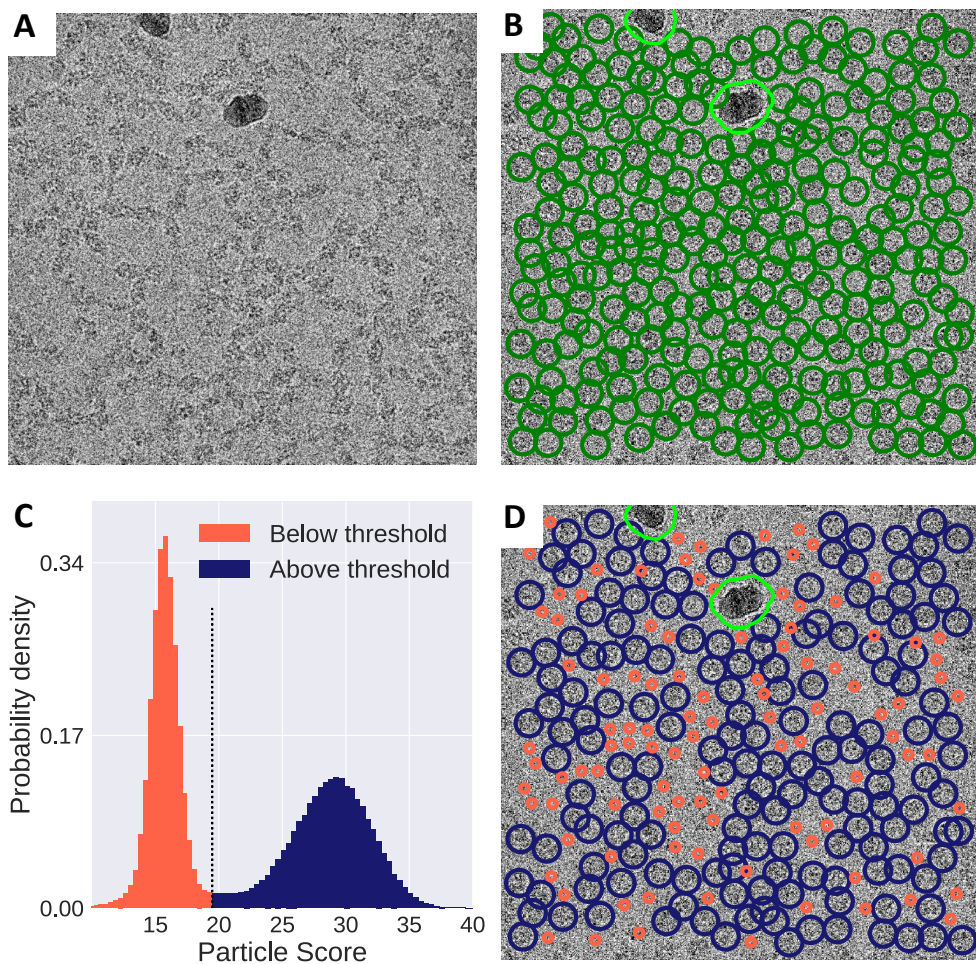

Figure 3. **Unsupervised strategy for particle picking and sorting.** **A)** Representative raw micrograph of beta-galactosidase from EMPIAR-10061. **B)** Automated exclusion of high-contrast areas (light green contours) and detection of particle locations (blue markers) based on expected particle dimensions. **C)** Bi-modal distribution of particle scores (across all particles in the dataset) obtained using the program cisTEM. Score values represent the quality-of-fit between experimental molecular images and the corresponding projection of an ab-initio 3D model. The two particle populations are estimated using a Gaussian mixture model and separated using the equal probability point (shown as vertical dashed line). Low-scoring particles falling below the threshold are indicated in orange and high-scoring particles are shown in blue. **D)** Mapping of results from the unsupervised particle sorting strategy showing the location of both populations of particles on the original micrograph. Only particles with score values above the threshold are used for further processing.

the subpopulation of particle images that are consistent with a single 3D model (particle sorting).

## 2.1 Strategies for particle picking

The goal of particle picking is to locate image features within micrographs that are consistent in appearance with the overall dimensions of the target structure. For this reason, many particle picking algorithms use mainly low-resolution information in order to detect features that are in the expected size range. It has been reported that the use of high-resolution information for the purpose of particle picking can potentially introduce bias and lead to problems during structure refinement [48]–[50]. Most particle picking methods are based on the detection of local extrema of *saliency* or *feature* maps computed from the image data. These maps capture information about the strength or likelihood of observing particles at individual pixel locations based on local contrast characteristics. Given a specific feature map, the detection sensitivity of these algorithms can be controlled by adjusting the threshold value used for detecting local maxima within the image. For example, higher thresholds will produce fewer false-positives but also more false-negatives (under-detection), while lower thresholds will produce fewer false-negatives but also increase the number of false-positives (over-detection).

## 2.2 Strategies for particle sorting

In addition to bona fide particles, real micrographs often contain other features that are in the same size range as the target of interest but instead correspond to ice contamination, broken, denatured or otherwise inconsistent particles with a consensus 3D reconstruction. This motivates the need to use particle sorting strategies in order to eliminate patches that do not correspond to true particle images. The majority of SPA packages perform particle sorting in an interactive fashion using one or multiple rounds of 2D and 3D image classification, **Figure 2**. Unlike many particle picking approaches that use contrast information from individual micrographs in order to detect the presence of particles, strategies for particle sorting integrate information from entire datasets, for example, by aligning and classifying large sets of particle images extracted from all micrographs in a dataset.

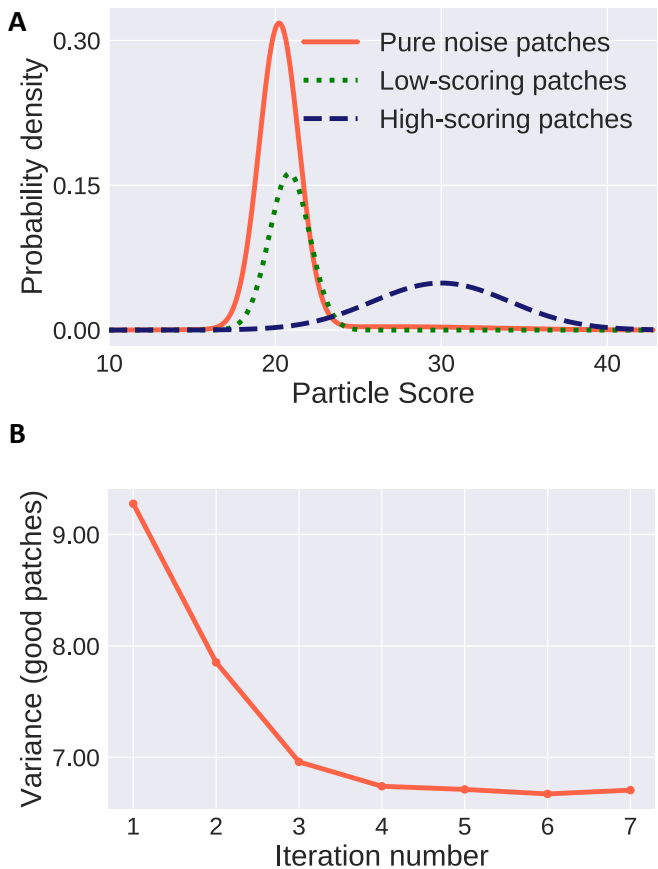

Figure 4: **Statistical analysis of particle scores assigned to individual image patches during 3D refinement**. **A)** Distribution of pure noise patches selected from the data (orange), as well as low-scoring particles (green) and high-scoring particles (blue) separated by the automated procedure. The noise peak overlaps with the low-scoring group of particles identified by the unsupervised selection procedure. **B)** Plot of the variance of the peak corresponding to the high-scoring particles (as a function of the refinement iteration), showing a narrowing of the distribution consistent with the predictions of our statistical model.

*2.3 Unsupervised particle sorting*

In this section, we describe our proposed two-step procedure for unsupervised particle picking and sorting. First, we detect particles by convolving each micrograph with a Gaussian kernel whose width matches the size of the target structure. Similar picking strategies are implemented in the most commonly used SPA packages in the field and any particle picking method may be used for this step. In order to minimize the number of false positives, we detect particles at the location of every local maxima of the convolution image with the Gaussian kernel. To deal with abnormally low or high contrast areas due to the presence of ice contamination, carbon, etc., we mask those regions out using simple image thresholding and mathematical morphology operations, **Figure 3A-B**. For the particle sorting stage, we assume an initial low-resolution 3D model of the target of interest is available, such as an ab-initio reconstruction produced by stochastic gradient descent [34]. We then use this model to determine the alignment parameters for each particle using the projection matching algorithm implemented in cisTEM [32]. During this process, each particle is assigned a *score* value that measures the quality-of-fit between the particle image and a projection of the model in the assigned orientation (higher score values represent a better match between the data and the model). The formula for the score is the normalized cross-correlation function computed in Fourier space over a frequency range selected by the user (after SNR-weighting), plus a term used to restrain alignment parameters [32]. A simple strategy to sort particle images would be to only keep the highest-scoring particles. This, however, requires the selection of an arbitrary cut-off value and depending on the actual fraction of good-vs-bad particles, may result in suboptimal separation of particles. By inspecting the histogram of score values across all particles in a dataset, we noticed a characteristic bimodal distribution consistent with the presence of two distinct particle populations. Our strategy for particle sorting consists of fitting a mixture of two Gaussians to the empirical score distribution and deriving a threshold to automatically separate the two populations, **Figure 3C-D**. The observation that particle scores exhibit this behaviour has been reported in the literature before [24], [51]–[53], but to our knowledge, no efforts have been made to use this as a criteria for automatically sorting particle images. To determine the threshold, we use the equal-probability point between the estimated normal modes of the distribution. Particles with score values above the threshold are extracted and used to generate a 3D map that in turn can be used as a reference for a new round of 3D refinement and particle sorting. By iterating this procedure until convergence, we are able to identify a consistent subset of particle images representing projections of a unique underlying 3D model.

### 3. Statistical Analysis

To understand the origin of the bimodal score distribution, we hypothesized that low-scoring patches may correspond to empty boxes (false-positives produced by the particle picking procedure) that could be modeled as pure Gaussian noise. To test the validity of this model, we conducted an experiment in which we deliberately selected empty patches from the EMPIAR-10025 dataset and compared their cisTEM score values against the bimodal distribution obtained using all the patches, **Figure 4A**. The score values obtained for the noise patches overlap with those of the low-scoring mode in the bimodal distribution, in agreement with our hypothesis. In what follows, we consider two populations of image patches: *good* patches which agree with our 3D reference, and pure *noise* or empty patches. The noise patches are modeled as pure Gaussian noise:

$$y_i^{\text{noise}} \sim \mathcal{N}(0, \sigma^2 I_{N^2}),$$

whereas good patches are modeled as:

$$y_j^{\text{good}} = P_j \mu + \epsilon_j.$$

Here, $\mu \in \mathbb{R}^{N^3}$ is the true (oracle) 3D density of the molecule, $\epsilon_j \sim \mathcal{N}(0, \sigma^2 I_{N^2})$ is Gaussian noise, and $P_j \in \mathbb{R}^{N^2 \times N^3}$ is the imaging operator of the j$^{\text{th}}$ patch. It takes a volume and performs a tomographic projection in the direction that exactly matches the particle's orientation, followed by application of the CTF, similar statistical models were used in [54], [55].

Assuming that the score values used in our experiments are proportional to the normalized cross-correlation (NCC) of each image patch with a projected reference, $\widetilde{\mu} \in \mathbb{R}^{N^3}$, assumed to be close to the true density $\mu$. For noise patches, the score is:

$$\text{NCC}(y_i^{\text{noise}}, \widetilde{P}_i \widetilde{\mu}) = \frac{\langle \epsilon_i, \widetilde{P}_i \widetilde{\mu} \rangle}{\|\epsilon_i\| \|\widetilde{P}_i \widetilde{\mu}\|},$$

where $\|\cdot\|$ denotes the $l_2$ norm, $\widetilde{P}_i$ is an imaging operator based on the estimated particle orientation. In the pure noise case, we model the estimated orientation as random. Note that $\frac{1}{\sigma^2}\|\epsilon_i\|^2$ is a chi-squared random variable with $N^2$ degrees of freedom. For large values of $N$, its value is concentrated around $N^2$. We approximate $\|\epsilon_i\| \approx \sigma N$, thus obtaining

$$\text{NCC}(y_i^{\text{noise}}, \widetilde{P}_i \widetilde{\mu}) = \frac{\langle \epsilon_i, \widetilde{P}_i \widetilde{\mu} \rangle}{\|\epsilon_i\| \|\widetilde{P}_i \widetilde{\mu}\|} \approx \frac{\langle \epsilon_i, \widetilde{u}_i \rangle}{\sigma N},$$

where $\widetilde{u}_i = \widetilde{P}_i \widetilde{\mu} / \|\widetilde{P}_i \widetilde{\mu}\|$ is a unit vector. The inner product of a $\mathcal{N}(0, \sigma^2 I)$ variable with any independently drawn unit vector is distributed as $\mathcal{N}(0, \sigma^2)$. Hence, under the approximation that $\|\epsilon_i\| \approx \sigma N$ we obtain the distribution of the noise patch scores:

$$\text{NCC}(y_i^{\text{noise}}, \widetilde{P}_i \widetilde{\mu}) \sim \mathcal{N}(0, 1/N^2).$$

We now consider the score for the *good* patches:

$$\text{NCC}(y_j^{\text{good}}, \widetilde{P}_j \widetilde{\mu}) = \frac{\langle P_j \mu + \epsilon_i, \widetilde{P}_j \widetilde{\mu} \rangle}{\|P_j \mu + \epsilon_j\| \|\widetilde{P}_j \widetilde{\mu}\|}.$$

In typical micrographs, the noise level is much greater than the signal, so we approximate the denominator as

$$\|P_j \mu + \epsilon_j\| \approx \|\epsilon_j\| \approx \sigma N.$$

With this approximation, we may write,

$$NCC(y_j^{good}, \tilde{P}_j \tilde{\mu}) \approx \frac{\langle P_j\mu, \tilde{P}_j\tilde{\mu} \rangle}{\sigma N \|\tilde{P}_j\tilde{\mu}\|} + \frac{\langle \epsilon_j, \widetilde{u}_j \rangle}{\sigma N}.$$

Decomposing the second term into three parts, we have:

$$\frac{\langle P_j\mu, \widetilde{P}_j\widetilde{\mu} \rangle}{\sigma N \|\widetilde{P}_j\widetilde{\mu}\|} = \frac{1}{\sigma N}(\langle \widetilde{P}_j\widetilde{\mu}, \widetilde{u}_j \rangle + \langle \widetilde{P}_j(\mu - \widetilde{\mu}), \widetilde{u}_j \rangle$$
$$+ \langle (P_j - \widetilde{P}_j)\mu, \widetilde{u}_j \rangle)$$
$$= \frac{1}{\sigma N}(\|\widetilde{P}_j\widetilde{\mu}\| + \langle \widetilde{P}_j(\mu - \widetilde{\mu}), \widetilde{u}_j \rangle$$
$$+ \langle (P_j - \widetilde{P}_j)\mu, \widetilde{u}_j \rangle).$$

We assume that the reference density is close to the oracle density $\mu$, in the sense that $\|\mu - \widetilde{\mu}\| = O(\delta_1)$ and the imaging operators satisfy $\|\widetilde{P}_j - P_j\| = O(\delta_2)$ in operator norm. In that case it can be shown that:

$$NCC(y_j^{good}, \tilde{P}_j \widetilde{\mu})$$
$$\approx \frac{\langle \epsilon_j, \widetilde{u}_j \rangle + \|\widetilde{P}_j\widetilde{\mu}\| + B_N O(\delta_1) + \|\mu\| O(\delta_2)}{\sigma N},$$

where $B_N$ is a resolution-dependent operator-norm bound on the estimated imaging operators. The first term is the same as the score of the pure noise patches. Since the second term is always positive, we conclude that, at least in the cases where the reference density is close to the true density and the orientations are estimated accurately, the mean score of the good patches should be higher than the mean score of the pure noise patches. Interestingly, for spherically shaped reference models, the second term in the equation above is nearly constant and thus does not contribute much to the variance. In that case, we expect to see a bimodal distribution of two Gaussian-like distributions of similar width. In contrast, for non-spherically shaped molecules, the second term is another source of variance for the good patch scores. In this case, we expect to see a wider peak for the good patches than for the noise patches. The third and fourth terms are related to errors in estimation and both contribute to the variance of the good patch scores. The third term is due to the error in the reference density whereas the fourth term is due to errors in the assignment of orientations. The magnitude of these terms should decrease as the accuracy of the 3D reference improves, leading to reduced variance of the good particle scores. This agrees with our experimental observations, **Figure 4B,** where we show that the peak of the good patches becomes progressively narrower as the number of refinement iterations increase.

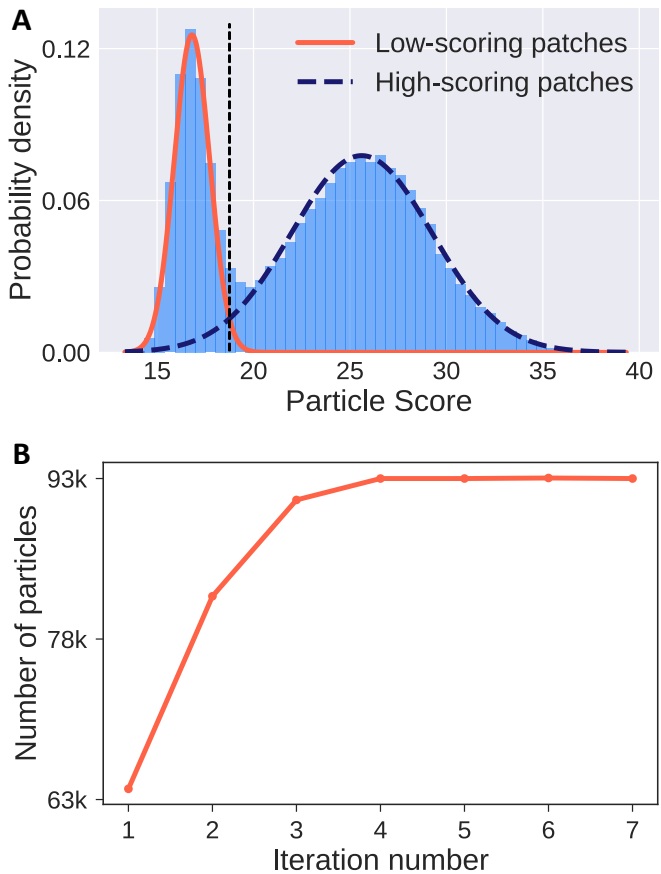

*Figure 5. **Threshold selection based on bimodal Gaussian mixture model**. **A)** Histogram of cisTEM score values assigned to individual particle images extracted from EMPIAR-10025 after convergence of the 3D refinement process. A bimodal Gaussian mixture model was fit to the data and the threshold selected as the equal-probability point between the two modes. **B)** Number of particles above the threshold selected after each refinement iteration. The number of high-scoring particles increases gradually as the resolution of the reconstruction improves and the orientation assignments become more accurate. After convergence, the number of particles in each group stabilizes and only residual changes in the assignment of particle orientations are observed.*

### 4. Results

To demonstrate the performance of our strategy for unsupervised particle selection, we processed data from the *Thermoplasma acidophilum 20S proteasome* available from the EMPIAR database (entry 10025). This dataset is composed of 196 movies collected on a Titan Krios TEM using a Gatan K2 Summit direct electron detector operated in super-resolution mode. Each movie has a total accumulated exposure of 53 e−/Å$^2$ fractionated into 38 frames. These images were used previously to produce a 2.8 Å resolution reconstruction available from the EMDB database (entry EMD-6287) [56].

We first aligned the raw frames and estimated the CTF parameters for each movie using UNBLUR [20] and CTFFIND4 [27], respectively. Particles were picked using a Gaussian disk of 60 Å in radius and we selected every local maxima from the convolution image in order to minimize the number of false positives. This produced a total of 171,622 particles. In order to speed up computations, particles were extracted using a binning factor of 4 and were assigned orientations in 3D using projection matching against an initial reference obtained using cisTEM's *ab-initio* 3D reconstruction routine [32]. A bimodal Gaussian mixture model was fit to the distribution of particle scores and the point of equal probability was selected as a threshold to separate the two populations, **Figure 5A**. Only particles with score values above the threshold were used to compute an updated 3D reconstruction which was in turn used as a reference to re-estimate the orientations of the original set of particles.

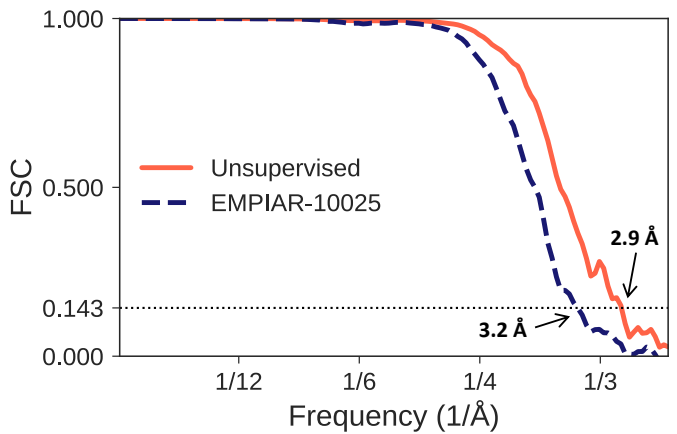

*Figure 6: **Improvement in map resolution obtained using automatically sorted particles vs. using the original set of particles selected by user supervision**. Fourier Shell Correlation (FSC) plots between half-maps obtained using the original set of 49,954 particles available from EMPIAR-10025 (blue) and the set of 96,633 particles obtained automatically using our sorting procedure (orange). Both sets of particles were re-extracted from the original micrographs and subjected to 3D auto-refinement using identical parameters and using the same ab-initio model as reference for alignment. The improvement in resolution measured using the 0.143-cutoff criteria is 0.3 Å, indicating that the unsupervised procedure produced almost double the number of valid particles.*

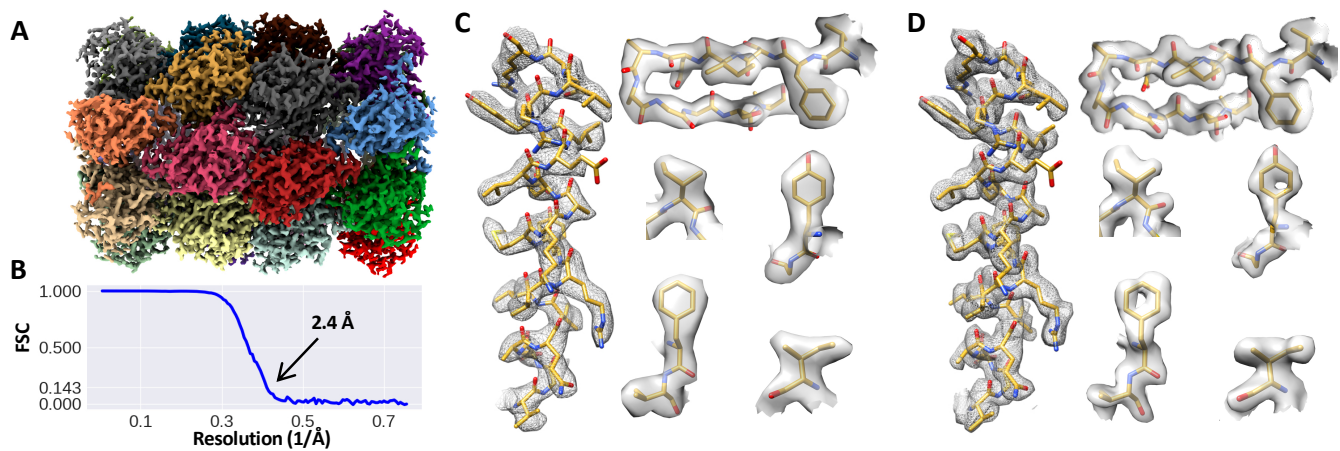

*Figure 7. Reconstruction of T20S proteasome obtained using unsupervised particle picking from EMPIAR-10025. A) Overview of high-resolution density map obtained from 93,633 particles selected automatically based on the bimodal distribution of cisTEM scores. B) Fourier Shell Correlation (FSC) plot between half-maps showing the estimated 2.4 Å resolution using the 0.143-cutoff criteria. C, D) Comparison of selected map regions between the original 2.8 Å map reported by Campbell et al. (C), and the 2.4 Å map we obtained using automatic particle sorting (D).*

This process was iterated for six additional rounds, and a different threshold was selected at each iteration based on the actual distribution of score values, resulting in a stable subset of 93,633 particles, **Figure 5B**. The number of particles selected at each iteration increased gradually, reflecting improvements in the accuracy of particle orientations due to the increased quality of the 3D reference and the finer rotational sampling done by the frequency marching refinement procedure. We note that the use of lower levels of particle binning (1x or 2x) would not affect the distribution of particle scores, because score evaluations are restricted to a frequency range that excludes resolutions beyond the Nyquist limit for 4x binning data. The only reason we used down-sampled images at this stage of processing was to speed up the search of particle orientations in 3D.

Our unsupervised sorting procedure produced almost twice as many particles as those used in the original study. To show that the additional particles produced by our approach indeed correspond to good particles, we downloaded particle coordinates used in [57] from the EMPIAR database and re-extracted the particles from the original micrographs (using the same box size and binning used to extract our automatically selected particles). Starting from the same initial model, we subjected both particle stacks to 3D auto-refinement (using identical parameters) and compared the FSC curves between half-maps for both reconstructions, **Figure 6**. The resolution of the map obtained using the automatically picked particles is higher than the one obtained using the original set of particles in [57], indicating that the additional particles produced by our approach indeed correspond to *good* particles.

The clean set of 93,633 particles was re-extracted from the original unbinned micrographs using a pixel size of 0.66 Å and subjected to 7 additional rounds of local 3D refinement as implemented in cisTEM (starting from the alignment parameters obtained using the binned data). We then applied two rounds of per-particle defocus refinement and drift-correction as described in [24], producing a final map at 2.4 Å resolution, **Figure 7A**. To assess the quality of the reconstruction, we computed the Fourier Shell Correlation (FSC) curve between half-maps, **Figure 7B**. We also compared our map against the original 2.8 Å map (EMDB-6287), by looking at the same map regions highlighted in [56], **Figure 7B-D**. Our map shows better resolved features, including the appearance of holes in aromatic rings and well-defined density for backbone carbonyls, features that were not observed in the original map and that are consistent with the resolution numbers reported by the FSC criteria. While the central point of our experiments was to show that particle sorting can be done in an unsupervised manner without the need to manually sort particles by 2D or 3D classification, our results also show that the automated sorting procedure is capable of producing state-of-the-art results in terms of map resolution. We note that several factors contributed to the improvement in resolution obtained over the original map [58], including the fact that we processed the data using a smaller pixel size (0.66 Å vs. 0.98 Å), almost twice as many particles were selected by our unsupervised sorting approach and used for the reconstruction (93,633 vs. 49,954 particles), and the fact that we applied strategies for per-particle CTF correction that were not used in the original study.

In summary, our results show that in the favourable case of homogeneous samples, such as the T20S proteasome, particle selection and sorting can be done in an automatic fashion, entirely bypassing the need to use supervised strategies for 2D/3D classification that are prone to user bias and can impact the reproducibility of the structure determination process.

## 5. Discussion

Despite recent advances in data analysis for single-particle cryo-EM, the overall structure determination process remains largely an inexact art that requires significant levels of user involvement in order to produce high-resolution structures. In particular, the sorting of particle images is one of the most time consuming and critical steps in the SPA pipeline that can often determine the success of the entire structure determination process. Here, we present a simple computational strategy to automate the process of particle sorting, allowing us to streamline and improve the consistency and reproducibility of SPA refinement routines.

The success of our strategy for unsupervised particle sorting relies on the availability of an accurate initial model that can be used as a reference to align the particle images. The only requirement for the initial model is that it should contain enough low-resolution features to allow the unambiguous alignment of the good particle images. This is necessary because our approach is based on analysing the distribution of score values of particle images measured against projections of the reference model. The robust *ab-initio* reconstruction models produced by stochastic gradient descent as implemented in recent SPA packages [32], [34], [59], generally produces models that are of sufficient quality to satisfy this requirement.

Another factor that may directly influence the performance of our particle sorting strategy is the molecular weight of the target of interest. The accuracy of image alignment routines used during 3D refinement is known to degrade for smaller molecular weight targets due to the reduction in image contrast [14], [60]. In practice, this means that for low molecular-weight complexes (~100kDa or less), the cross-correlation based scoring function used during projection matching in cisTEM may fail to discriminate between different particle orientations and between empty patches and good particles, and the histogram of score values will no longer have a bimodal distribution. In this case, our approach will fail to separate the different particle populations and alternative supervised procedures for particle sorting would need to be used instead. Finally, the presence of conformational heterogeneity within the sample is another factor to consider regarding the performance of our strategy for particle sorting. In general, while our approach will be unable to distinguish between particles from different conformations, it may still be effective at eliminating junk particles or false positives produced by the particle picking procedure. In this case, application of our strategy for particle sorting to heterogeneous samples will provide a *clean* set of particles that can later be subjected to supervised 3D classification in order to separate the different conformations present in the sample.

In principle, the sensitivity of our method to small molecular-weight and heterogeneous targets may be mitigated by computing the particle scores using a Bayesian approach [61]. In this framework, rather than choosing the orientation that best fits the experimental image, a marginalized likelihood is computed over all possible orientations. In addition, different conformational states can be incorporated into the likelihood function and thus integrated naturally into the framework, as implemented in several SPA packages [32], [34], [40]. One drawback of this method, however, is the elevated computational cost required to evaluate the marginalized likelihood over all possible rotations and translations for each experimental image. Under certain conditions, however, the computational load may be alleviated by employing multi-scale methods or advanced approximation techniques [62].

## 6. Conclusion

Recent technological advances in detector technology and image analysis have transformed cryo-EM into a powerful technique capable of studying the high-resolution structure of proteins and protein complexes. The availability of strategies for automated data collection and the introduction of larger and faster detectors, have dramatically improved the throughput of data production exacerbating the need to streamline the cryo-EM structure determination pipeline. Particle selection and sorting is one the most time-consuming and critical steps in the SPA workflow that still relies on subjective user input in order to successfully identify homogeneous populations of particles that can be used for high-resolution refinement.

In an effort to address this problem, we presented an unsupervised strategy for particle sorting that lessens the burden on users while at the same time speeds-up the structure determination process and improves its reproducibility and consistency. We showed that in the favorable case of large and homogeneous complexes like the T20S proteasome, particle sorting can be executed in an unsupervised manner yielding state-of-the-art reconstructions that compare favorably with those obtained by expert users. Whether the strategy proposed here can be applied to a wider and more challenging class of samples, including complexes of smaller molecular weight, lower symmetry or that are more heterogeneous, will be the subject of future studies. This and other advances in automation in single-particle cryo-EM will eventually put the technique on a path of establishing itself as a mainstream technique for high-resolution structure determination with the potential of studying the structure of important biomolecules in high-throughput mode of operation.

## Acknowledgements

This study utilized the computational resources offered by Duke Research Computing (http://rc.duke.edu) at Duke

University. We thank M. DeLong, C. Kneifel, M. Newton, V. Orlikowski, T. Milledge, and D. Lane from the Duke Office of Information Technology and Research Computing for providing assistance with setting up and maintaining the computing environment.

**References**

[1] Kühlbrandt, W., "Cryo-EM enters a new era," *Elife*, vol. 3, Aug. 2014.

[2] Egelman, E. H., "The Current Revolution in Cryo-EM.," *Biophys. J.*, vol. 110, no. 5, pp. 1008–12, Mar. 2016.

[3] Velankar, S., G. van Ginkel, Y. Alhroub, G. M. Battle, J. M. Berrisford, M. J. Conroy, J. M. Dana, S. P. Gore, A. Gutmanas, P. Haslam, P. M. S. Hendrickx, I. Lagerstedt, S. Mir, M. A. Fernandez Montecelo, A. Mukhopadhyay, T. J. Oldfield, A. Patwardhan, E. Sanz-García, S. Sen, R. A. Slowley, M. E. Wainwright, M. S. Deshpande, A. Iudin, G. Sahni, J. Salavert Torres, M. Hirshberg, L. Mak, N. Nadzirin, D. R. Armstrong, A. R. Clark, O. S. Smart, P. K. Korir and G. J. Kleywegt, "PDBe: improved accessibility of macromolecular structure data from PDB and EMDB.," *Nucleic Acids Res.*, vol. 44, no. D1, pp. 385–95, 2016.

[4] Guo, T., A. Bartesaghi, H. Yang, V. Falconieri, P. Rao, A. Merk, E. Eng, A. Raczkowski, T. Fox, L. Earl, D. Patel and S. Subramaniam, "Cryo-EM Structures Reveal Mechanism and Inhibition of DNA Targeting by a CRISPR-Cas Surveillance Complex," *Cell*, vol. 171, no. 2, pp. 414-426.e12, 2017.

[5] Banerjee, S., A. Bartesaghi, A. Merk, P. Rao, S. Bulfer, Y. Yan, N. Green, B. Mroczkowski, R. Neitz, P. Wipf, V. Falconieri, R. Deshaies, J. Milne, D. Huryn, M. Arkin and S. Subramaniam, "2.3 A resolution cryo-EM structure of human p97 and mechanism of allosteric inhibition," *Science (80-. ).*, vol. 351, no. 6275, pp. 871–875, 2016.

[6] Matthies, D., C. Bae, G. E. Toombes, T. Fox, A. Bartesaghi, S. Subramaniam and K. J. Swartz, "Single-particle cryo-EM structure of a voltage-activated potassium channel in lipid nanodiscs," *Elife*, vol. 7, 2018.

[7] Liao, M., E. Cao, D. Julius and Y. Cheng, "Structure of the TRPV1 ion channel determined by electron cryo-microscopy," *Nature*, vol. 504, no. 7478, pp. 107–112, 2013.

[8] Matthies, D., O. Dalmas, M. Borgnia, P. Dominik, A. Merk, P. Rao, B. Reddy, S. Islam, A. Bartesaghi, E. Perozo and S. Subramaniam, "Cryo-EM Structures of the Magnesium Channel CorA Reveal Symmetry Break upon Gating," *Cell*, vol. 164, no. 4, pp. 747–756, 2016.

[9] Meyerson, J. R., J. Kumar, S. Chittori, P. Rao, J. Pierson, A. Bartesaghi, M. L. Mayer and S. Subramaniam, "Structural mechanism of glutamate receptor activation and desensitization," *Nature*, vol. advance on, 2014.

[10] Kang, Y., O. Kuybeda, P. de Waal, S. Mukherjee, N. Van Eps, P. Dutka, X. Zhou, A. Bartesaghi, S. Erramilli, T. Morizumi, X. Gu, Y. Yin, P. Liu, Y. Jiang, X. Meng, G. Zhao, K. Melcher, O. Ernst, A. Kossiakoff, S. Subramaniam and H. Xu, "Cryo-EM structure of human rhodopsin bound to an inhibitory G protein," *Nature*, vol. 558, pp. 553–558, 2018.

[11] García-Nafría, J., Y. Lee, X. Bai, B. Carpenter and C. G. Tate, "Cryo-EM structure of the adenosine A2A receptor coupled to an engineered heterotrimeric G protein," *Elife*, vol. 7, May 2018.

[12] García-Nafría, J., R. Nehmé, P. C. Edwards and C. G. Tate, "Cryo-EM structure of the serotonin 5-HT1B receptor coupled to heterotrimeric Go," *Nature*, vol. 558, no. 7711, pp. 620–623, Jun. 2018.

[13] Liang, Y.-L., M. Khoshouei, M. Radjainia, Y. Zhang, A. Glukhova, J. Tarrasch, D. M. Thal, S. G. B. Furness, G. Christopoulos, T. Coudrat, R. Danev, W. Baumeister, L. J. Miller, A. Christopoulos, B. K. Kobilka, D. Wootten, G. Skiniotis and P. M. Sexton, "Phase-plate cryo-EM structure of a class B GPCR-G-protein complex.," *Nature*, vol. 546, no. 7656, pp. 118–123, 2017.

[14] Merk, A., A. Bartesaghi, S. Banerjee, V. Falconieri, P. Rao, M. Davis, R. Pragani, M. Boxer, L. Earl, J. Milne and S. Subramaniam, "Breaking Cryo-EM Resolution Barriers to Facilitate Drug Discovery," *Cell*, vol. 165, no. 7, pp. 1698–1707, 2016.

[15] Bartesaghi, A., A. Merk, S. Banerjee, D. Matthies, X. Wu, J. L. S. Milne and S. Subramaniam, "2.2 å resolution cryo-EM structure of β-galactosidase in complex with a cell-permeant inhibitor," *Science (80-. ).*, vol. 348, no. 6239, pp. 1147–1151, 2015.

[16] Borgnia, M., S. Banerjee, A. Merk, D. Matthies, A. Bartesaghi, P. Rao, J. Pierson, L. Earl, V. Falconieri, S. Subramaniam and J. Milne, "Using Cryo-EM to Map Small Ligands on Dynamic Metabolic Enzymes: Studies with Glutamate Dehydrogenase," *Mol. Pharmacol.*, vol. 89, no. 6, pp. 645–651, 2016.

[17] Brilot, A. F., J. Z. Chen, A. Cheng, J. Pan, S. C. Harrison, C. S. Potter, B. Carragher, R. Henderson and N. Grigorieff, "Beam-induced motion of vitrified specimen on holey carbon film.," *J. Struct. Biol.*, vol. 177, no. 3, pp. 630–7, 2012.

[18] Bartesaghi, A., D. Matthies, S. Banerjee, A. Merk and S. Subramaniam, "Structure of -galactosidase at 3.2-A resolution obtained by cryo-electron microscopy," *Proc. Natl. Acad. Sci.*, vol. 111, no. 32, pp. 11709–11714, 2014.

[19] Kremer, J. R., D. N. Mastronarde and J. R. McIntosh, "Computer visualization of three-dimensional image data using IMOD.," *J. Struct. Biol.*, vol. 116, no. 1, pp. 71–76, 1996.

[20] Grant, T. and N. Grigorieff, "Measuring the optimal exposure for single particle cryo-EM using a 2.6 Å reconstruction of rotavirus VP6," *Elife*, vol. 4, 2015.

[21] Li, X., P. Mooney, S. Zheng, C. R. Booth, M. B.

Braunfeld, S. Gubbens, D. A. Agard and Y. Cheng, "Electron counting and beam-induced motion correction enable near-atomic-resolution single-particle cryo-EM.," *Nat. Methods*, vol. 10, no. 6, pp. 584–90, 2013.

[22] Zheng, S. Q., E. Palovcak, J.-P. Armache, K. A. Verba, Y. Cheng and D. A. Agard, "MotionCor2: anisotropic correction of beam-induced motion for improved cryo-electron microscopy," *Nat. Methods*, vol. 14, no. 4, pp. 331–332, 2017.

[23] Abrishami, V., J. Vargas, X. Li, Y. Cheng, R. Marabini, C. Ó. S. Sorzano and J. M. Carazo, "Alignment of direct detection device micrographs using a robust Optical Flow approach.," *J. Struct. Biol.*, vol. 189, no. 3, pp. 163–76, 2015.

[24] Bartesaghi, A., C. Aguerrebere, V. Falconieri, S. Banerjee, L. A. Earl, X. Zhu, N. Grigorieff, J. L. S. Milne, G. Sapiro, X. Wu and S. Subramaniam, "Atomic Resolution Cryo-EM Structure of β-Galactosidase," *Structure*, vol. 26, no. 6, pp. 848-856.e3, 2018.

[25] Rubinstein, J. L. and M. A. Brubaker, "Alignment of cryo-EM movies of individual particles by optimization of image translations," *J. Struct. Biol.*, vol. 192, no. 2, pp. 188–195, 2015.

[26] Scheres, S. H., "Beam-induced motion correction for sub-megadalton cryo-EM particles.," *Elife*, vol. 3, p. e03665, 2014.

[27] Rohou, A. and N. Grigorieff, "CTFFIND4: Fast and accurate defocus estimation from electron micrographs," *J. Struct. Biol.*, vol. 192, no. 2, pp. 216–221, 2015.

[28] Zhang, K., "Gctf: Real-time CTF determination and correction," *J. Struct. Biol.*, vol. 193, no. 1, pp. 1–12, 2016.

[29] Wagner, T., F. Merino, M. Stabrin, T. Moriya, C. Gatsogiannis and S. Raunser, "SPHIRE-crYOLO: A fast and well-centering automated particle picker for cryo-EM," *bioRxiv*, p. 356584, Jun. 2018.

[30] Bepler, T., A. Morin, J. Brasch, L. Shapiro, A. J. Noble and B. Berger, "Positive-unlabeled convolutional neural networks for particle picking in cryo-electron micrographs," Mar. 2018.

[31] Wang, F., H. Gong, G. Liu, M. Li, C. Yan, T. Xia, X. Li and J. Zeng, "DeepPicker: A deep learning approach for fully automated particle picking in cryo-EM," *J. Struct. Biol.*, vol. 195, no. 3, pp. 325–336, 2016.

[32] Grant, T., A. Rohou and N. Grigorieff, "cisTEM, user-friendly software for single-particle image processing," *Elife*, vol. 7, 2018.

[33] Heimowitz, A., J. Andén and A. Singer, "APPLE picker: Automatic particle picking, a low-effort cryo-EM framework," *J. Struct. Biol.*, vol. 204, no. 2, pp. 215–227, Nov. 2018.

[34] Punjani, A., J. L. Rubinstein, D. J. Fleet and M. A. Brubaker, "cryoSPARC: algorithms for rapid unsupervised cryo-EM structure determination," *Nat. Methods*, vol. 14, no. 3, pp. 290–296, 2017.

[35] Zivanov, J., T. Nakane, B. O. Forsberg, D. Kimanius, W. J. Hagen, E. Lindahl and S. H. Scheres, "New tools for automated high-resolution cryo-EM structure determination in RELION-3," *Elife*, vol. 7, 2018.

[36] Moriya, T., M. Saur, M. Stabrin, F. Merino, H. Voicu, Z. Huang, P. A. Penczek, S. Raunser and C. Gatsogiannis, "High-resolution Single Particle Analysis from Electron Cryo-microscopy Images Using SPHIRE.," *J. Vis. Exp.*, no. 123, 2017.

[37] Tang, G., L. Peng, P. R. Baldwin, D. S. Mann, W. Jiang, I. Rees and S. J. Ludtke, "EMAN2: an extensible image processing suite for electron microscopy.," *J. Struct. Biol.*, vol. 157, no. 1, pp. 38–46, 2007.

[38] Greenberg, I. and Y. Shkolnisky, "Common lines modeling for reference free Ab-initio reconstruction in cryo-EM," *J. Struct. Biol.*, vol. 200, no. 2, pp. 106–117, Nov. 2017.

[39] Levin, E., T. Bendory, N. Boumal, J. Kileel and A. Singer, "3D ab initio modeling in cryo-EM by autocorrelation analysis," in *2018 IEEE 15th International Symposium on Biomedical Imaging (ISBI 2018)*, 2018, pp. 1569–1573.

[40] Scheres, S. H. W., "RELION: Implementation of a Bayesian approach to cryo-EM structure determination," *J. Struct. Biol.*, vol. 180, no. 3, pp. 519–530, 2012.

[41] Wang, R. Y.-R., Y. Song, B. A. Barad, Y. Cheng, J. S. Fraser and F. DiMaio, "Automated structure refinement of macromolecular assemblies from cryo-EM maps using Rosetta," *Elife*, vol. 5, 2016.

[42] Brown, A., F. Long, R. A. Nicholls, J. Toots, P. Emsley and G. Murshudov, "Tools for macromolecular model building and refinement into electron cryo-microscopy reconstructions," *Acta Crystallogr. Sect. D Biol. Crystallogr.*, vol. 71, no. 1, pp. 136–153, 2015.

[43] Sorzano, C. O. S., A. Jiménez, J. Mota, J. L. Vilas, D. Maluenda, M. Martínez, E. Ramírez-Aportela, T. Majtner, J. Segura, R. Sánchez-García, Y. Rancel, L. del Caño, P. Conesa, R. Melero, S. Jonic, J. Vargas, F. Cazals, Z. Freyberg, J. Krieger, I. Bahar, R. Marabini, J. M. Carazo and IUCr, "Survey of the analysis of continuous conformational variability of biological macromolecules by electron microscopy," *Acta Crystallogr. Sect. F Struct. Biol. Commun.*, vol. 75, no. 1, pp. 19–32, Jan. 2019.

[44] Bendory, T., A. Bartesaghi and A. Singer, "Single-particle cryo-electron microscopy: Mathematical theory, computational challenges, and opportunities," Aug. 2019.

[45] Carragher, B., N. Kisseberth, D. Kriegman, R. A. Milligan, C. S. Potter, J. Pulokas and A. Reilein, "Leginon: An Automated System for Acquisition of Images from Vitreous Ice Specimens," *J. Struct. Biol.*, vol. 132, no. 1, pp. 33–45, 2000.

[46] Mastronarde, D. N., "Automated electron microscope

tomography using robust prediction of specimen movements," *J. Struct. Biol.*, vol. 152, no. 1, pp. 36–51, 2005.

[47] van Heel, M., G. Harauz, E. V. Orlova, R. Schmidt and M. Schatz, "A New Generation of the IMAGIC Image Processing System," *J. Struct. Biol.*, vol. 116, no. 1, pp. 17–24, 1996.

[48] Henderson, R., "Avoiding the pitfalls of single particle cryo-electron microscopy: Einstein from noise.," *Proc. Natl. Acad. Sci. U. S. A.*, vol. 110, no. 45, pp. 18037–41, Nov. 2013.

[49] van Heel, M., "Finding trimeric HIV-1 envelope glycoproteins in random noise.," *Proc. Natl. Acad. Sci. U. S. A.*, vol. 110, no. 45, pp. E4175-7, Nov. 2013.

[50] Subramaniam, S., "Structure of trimeric HIV-1 envelope glycoproteins.," *Proc. Natl. Acad. Sci. U. S. A.*, vol. 110, no. 45, pp. E4172-4, Nov. 2013.

[51] Sorzano, C. O. S., J. Vargas, J. M. de la Rosa-Trevín, A. Jiménez, D. Maluenda, R. Melero, M. Martínez, E. Ramírez-Aportela, P. Conesa, J. L. Vilas, R. Marabini and J. M. Carazo, "A new algorithm for high-resolution reconstruction of single particles by electron microscopy," *J. Struct. Biol.*, vol. 204, no. 2, pp. 329–337, 2018.

[52] Zhang, X., E. Settembre, C. Xu, P. R. Dormitzer, R. Bellamy, S. C. Harrison and N. Grigorieff, "Near-atomic resolution using electron cryomicroscopy and single-particle reconstruction," *Proc. Natl. Acad. Sci.*, vol. 105, no. 6, pp. 1867–1872, 2008.

[53] Afanasyev, P., C. Seer-Linnemayr, R. B. G. Ravelli, R. Matadeen, S. De Carlo, B. Alewijnse, R. V. Portugal, N. S. Pannu, M. Schatz and M. van Heel, "Single-particle cryo-EM using alignment by classification (ABC): the structure of *Lumbricus terrestris* haemoglobin," *IUCrJ*, vol. 4, no. 5, pp. 678–694, Sep. 2017.

[54] Singer, A., "Angular synchronization by eigenvectors and semidefinite programming," *Appl. Comput. Harmon. Anal.*, vol. 30, no. 1, pp. 20–36, Jan. 2011.

[55] Singer, A. and Y. Shkolnisky, "Three-Dimensional Structure Determination from Common Lines in Cryo-EM by Eigenvectors and Semidefinite Programming," *SIAM J. Imaging Sci.*, vol. 4, no. 2, pp. 543–572, Jan. 2011.

[56] Campbell, M. G., D. Veesler, A. Cheng, C. S. Potter and B. Carragher, "2.8 Å resolution reconstruction of the Thermoplasma acidophilum 20S proteasome using cryo-electron microscopy," *Elife*, vol. 4, 2015.

[57] Campbell, M. G., D. Veesler, A. Cheng, C. S. Potter and B. Carragher, "2.8 Å resolution reconstruction of the Thermoplasma acidophilum 20S proteasome using cryo-electron microscopy," *Elife*, vol. 4, 2015.

[58] Rawson, S., M. G. Iadanza, N. A. Ranson and S. P. Muench, "Methods to account for movement and flexibility in cryo-EM data processing," *Methods*, vol. 100, pp. 35–41, 2016.

[59] Zivanov, J., T. Nakane, B. O. Forsberg, D. Kimanius, W. J. Hagen, E. Lindahl and S. H. Scheres, "New tools for automated high-resolution cryo-EM structure determination in RELION-3," *Elife*, vol. 7, 2018.

[60] Rosenthal, P. B. and R. Henderson, "Optimal Determination of Particle Orientation, Absolute Hand, and Contrast Loss in Single-particle Electron Cryomicroscopy," *J. Mol. Biol.*, vol. 333, no. 4, pp. 721–745, 2003.

[61] Sigworth, F. J., "A Maximum-Likelihood Approach to Single-Particle Image Refinement," *J. Struct. Biol.*, vol. 122, no. 3, pp. 328–339, 1998.

[62] Rangan, A., M. Spivak, J. Andén and A. Barnett, "Factorization of the translation kernel for fast rigid image alignment," May 2019.